\documentclass[final,5p,times,twocolumn]{elsarticle}
\usepackage[hidelinks]{hyperref}

\usepackage{amssymb}
\usepackage{amsthm}
\usepackage{amsmath}

\usepackage{booktabs}

\journal{Neurocomputing}

\begin{document}

\begin{frontmatter}



\title{Addressing Heterophily in Node Classification with Graph Echo State Networks}


\author[unipi]{Alessio Micheli}
\ead{micheli@di.unipi.it}

\author[unipi]{Domenico Tortorella\corref{cor1}}
\ead{domenico.tortorella@phd.unipi.it}

\affiliation[unipi]{organization={Department of Computer Science, University of Pisa},
  addressline={Largo Bruno Pontecorvo 3}, 
  city={Pisa},
  postcode={56127}, 
  country={Italy}}

\cortext[cor1]{Corresponding author}

\begin{abstract}
Node classification tasks on graphs are addressed via fully-trained deep message-passing models that learn a hierarchy of node representations via multiple aggregations of a node's neighbourhood.
While effective on graphs that exhibit a high ratio of intra-class edges, this approach poses challenges in the opposite case, i.e. heterophily, where nodes belonging to the same class are usually further apart.
In graphs with a high degree of heterophily, the smoothed representations based on close neighbours computed by convolutional models are no longer effective.
So far, architectural variations in message-passing models to reduce excessive smoothing or rewiring the input graph to improve longer-range message passing have been proposed.
In this paper, we address the challenges of heterophilic graphs with Graph Echo State Network (GESN) for node classification.
GESN is a reservoir computing model for graphs, where node embeddings are recursively computed by an untrained message-passing function.
Our experiments show that reservoir models are able to achieve better or comparable accuracy with respect to most fully trained deep models that implement \emph{ad hoc} variations in the architectural bias or perform rewiring as a preprocessing step on the input graph, with an improvement in terms of efficiency/accuracy trade-off.
Furthermore, our analysis shows that GESN is able to effectively encode the structural relationships of a graph node, by showing a correlation between iterations of the recursive embedding function and the distribution of shortest paths in a graph.
\end{abstract}

\begin{keyword}
Graph Neural Networks \sep Node Classification \sep Heterophily \sep Reservoir Computing
\end{keyword}

\end{frontmatter}


\section{Introduction}
\label{sec:intro}

Relations between entities, such as paper citations, links between web pages, user interactions on social networks, or bonds between atoms in molecules, can be best represented by graphs.
Tasks on this class of data require learning representations that encode not only an entity, but also the complex context due to its relationship network.
Since the introduction of pioneering models such as \textsl{Neural Network for Graphs} \citep{Micheli2009} and \textsl{Graph Neural Network} \citep{Scarselli2009}, a plethora of neural models have been proposed to solve graph-, edge-, and node-level tasks \citep{Bacciu2020,Wu2021,Battaglia2018}.
Most of these models share an architecture structured in layers that perform local aggregations of node features, e.g. graph convolution networks \citep{Micheli2009,Duvenaud2015,Atwood2016,Kipf2017}.
More generally, this mechanism is known as `message-passing', and has allowed the adaptive processing of graph data, i.e. the same model can generalize to different connectivity and different graphs without changes in the neural model structure.

The architectural bias of performing progressive aggregations of neighbouring node representations has proved particularly effective on graphs that exhibit an high degree of \emph{homophily}, e.g. social networks where people tend to establish relationships with their similars \cite{McPherson2001}.
Convolutional models tend to produce similar embeddings for close nodes, thanks to a \emph{smoothing} effect that allows to filter out noise and reconstruct missing information \cite{Keriven2022}.
However, many real-world graphs depart from this setting, and present a large number of inter-class edges, thus exhibiting \emph{heterophily}.
In this case, representations and predictions that rely chiefly on a node's immediate neighbours can be misleading, since nodes belonging to the same class are generally further apart.
Hence, a model may need the ability to learn long-range relationships between nodes, which in message-passing models can be achieved by stacking multiple aggregation layers to enlarge the receptive field.
However, on (semi-)supervised node classification tasks (i.e., classifying nodes by learning from a labelled subset of the graph) accuracy has been shown to decay as the number of layers increases in deep graph convolutional networks \cite{Li2018}, due to the collapse of node representations \cite{Oono2020}; this phenomenon is called \emph{over-smoothing}.
This problem is thus exacerbated in the case of heterophilic graphs \cite{Yan2022}.
In general, the inability to extract meaningful features in deeper layers for tasks that require discovering long-range relationships between nodes is called \emph{under-reaching}.
Alon \& Yahav \cite{Alon2021} maintain that one of its causes is \emph{over-squashing}: the problem of encoding an exponentially growing receptive field \cite{Micheli2009} in a fixed-size node embedding dimension.
In order to address the disadvantageous setting caused by the diverse challenges that message-passing models have to face, different architectural variations in convolution-based model have been proposed, such as expanding the radius of neighbourhood aggregations or introducing skip connections to exploit the full hierarchy of representations \cite{Zhu2020}.
Some models even propose abandoning message-passing altogether \cite{Lim2021}.
Apart from changing the models' architectural bias, another solution proposed is altering the connectivity (rewiring) of the input graph, in order to ease message-passing by removing structural `bottlenecks' that prevent long-range relationships to be learnt \cite{Topping2022}.

Graph Echo State Network (GESN) \cite{Gallicchio2010} is an efficient model within the reservoir computing (RC) paradigm.
In RC, input data is encoded via a randomly-initialized reservoir, while only a linear readout requires training \cite{Nakajima2021}.
GESN has already been successfully applied to graph-level classification tasks \cite{Gallicchio2020}.
We extended this model to node-level tasks.
In this paper, we build upon the preliminary results on its first application to node classification tasks \cite{Tortorella2022esann}, focusing in particular on the efficacy in tackling heterophilic graphs.
As our analysis will show, GESN is able to effectively represent the structural relationships of a node by going beyond the previously-established stability constraints required for graph-level tasks \cite{Tortorella2022}.
The reservoir computing paradigm allows us to decouple the challenges intrinsic to the tasks from those inherent to training deep convolutional models.
In particular, we will observe that our model does not suffer from the over-smoothing effects that plague fully-trained deep convolutional models, despite performing many more message-passing steps.

The remaining of this paper is organized as follows.
In section \ref{sec:background}, we present the general task of node classification, along with its challenges and the solutions proposed so far in literature.
We then introduce GESN for node classification in section \ref{sec:gesn}.
In section \ref{sec:experiments}, our model is evaluated on 19 node classification tasks ranging from medium- to large-scale graphs with different degrees of heterophily, while its accuracy and efficiency is compared against a broad class of fully-trained models.
The factors contributing to the effectiveness of GESN are analysed in section \ref{sec:analysis}.
Finally, we draw our conclusions in section \ref{sec:conclusion}.

\section{Node Classification}
\label{sec:background}

Let $\mathcal{G} = (\mathcal{V}, \mathcal{E})$ denote a graph with nodes $v \in \mathcal{V}$ and edges $(v,u) \in \mathcal{E}$, having node feature vectors $\mathbf{x}_v \in \mathbb{R}^X$ for each node $v \in \mathcal{V}$.
We denote by $\mathcal{N}_r(v)$ the $r$-neighbourhood of the ego node $v$, i.e. the set of nodes that can reach $v$ via a path within $r$ hops, and by $\mathbf{A}$ the graph adjacency matrix.
The goal of a (semi-)supervised node classification task is to learn a model from a subset $\mathcal{V}_{\mathrm{train}} \subset \mathcal{V}$ of graph nodes with known target labels $\{(\mathbf{x}_v, y_v)\}_{v \in \mathcal{V}_{\mathrm{train}}}$, in order to infer the node labels $y_v \in \{1, ..., C\}$ for the remaining nodes $\mathcal{V} \setminus \mathcal{V}_{\mathrm{train}}$ using the network structure and input features $\mathbf{x}_v$.

\paragraph{Message-passing models}

Neural network models that are able to process the input graph adaptively to learn node embeddings are based on the message-passing architecture.
Most common graph models are structured in $L$ layers, where each layer learns an embedding for each node based on an increasingly large receptive field \cite{Micheli2009}.
These layers $\ell \geq 1$ can be formalized as \cite{Xu2019}
\begin{equation}\label{eq:gnn-layer}
	\mathbf{h}_v^{(\ell)} = \textsc{combine}\left(\mathbf{h}_v^{(\ell-1)}, \textsc{aggregate}(\{\mathbf{h}_{u}^{(\ell-1)} : u \in \mathcal{N}_1(v)\})\right),
\end{equation}
where node embeddings $\mathbf{h}_v^{(\ell)} \in \mathbb{R}^H$ of layer $\ell$ are obtained by aggregating the previous embeddings $\mathbf{h}_{v'}^{(\ell-1)}$ of node $v$'s $1$-hop neighbours via $\textsc{aggregate}(\cdot)$, and then combined with the node's previous embeddings $\mathbf{h}_v^{(\ell-1)}$ via $\textsc{combine}(\cdot)$.
The first layer $\ell = 1$ either receives in input the original node features, $\mathbf{h}_v^{(0)} = \mathbf{x}_v$, or their embeddings, e.g.  $\mathbf{h}_v^{(0)} = \mathrm{MLP}(\mathbf{x}_v)$.
The final layer $L$ either directly predicts the one-hot encoding of target label $y_v$, or is followed by an MLP that serves this purpose.
The whole model is trained end-to-end by typically minimizing the cross-entropy loss.

The choice of functions in \eqref{eq:gnn-layer} determines the architectural bias of the model.
For example, GCN \cite{Kipf2017} layers are defined as
\begin{equation}\label{eq:gcn}
	\mathbf{h}_v^{(\ell)} = \mathrm{relu}\left(\sum_{u \in \mathcal{N}_1(v)} \mathbf{\hat{A}}_{v,u} \mathbf{W}^{(\ell)} \mathbf{h}_{u}^{(\ell-1)}\right),
\end{equation}
where $\mathbf{\hat{A}}$ is the normalized adjacency matrix, $\mathbf{W}^{(\ell)}$ are learnable weights, and input node features $\mathbf{x}_v \in \mathbb{R}^X$ in layer $\ell = 1$.
The local aggregation of neighbouring node representations of \eqref{eq:gcn} is called graph \emph{convolution}, in analogy with the 2D convolution operation in neural networks for images or 1D convolution for time series.
Models of this class include GraphSAGE \cite{Hamilton2017}, which averages neighbouring nodes' representations, and GAT \cite{Velickovic2018}, which performs a weighted average of neighbours via learned attention scores.
For the latter model, the $\textsc{aggregate}$ and $\textsc{combine}$ functions in \eqref{eq:gnn-layer} are actually combined into a single step, as attention scores are computed based on both ego and neighbour nodes embeddings.

\paragraph{Issues and challenges}

\begin{figure}
\centering
\includegraphics[scale=.25]{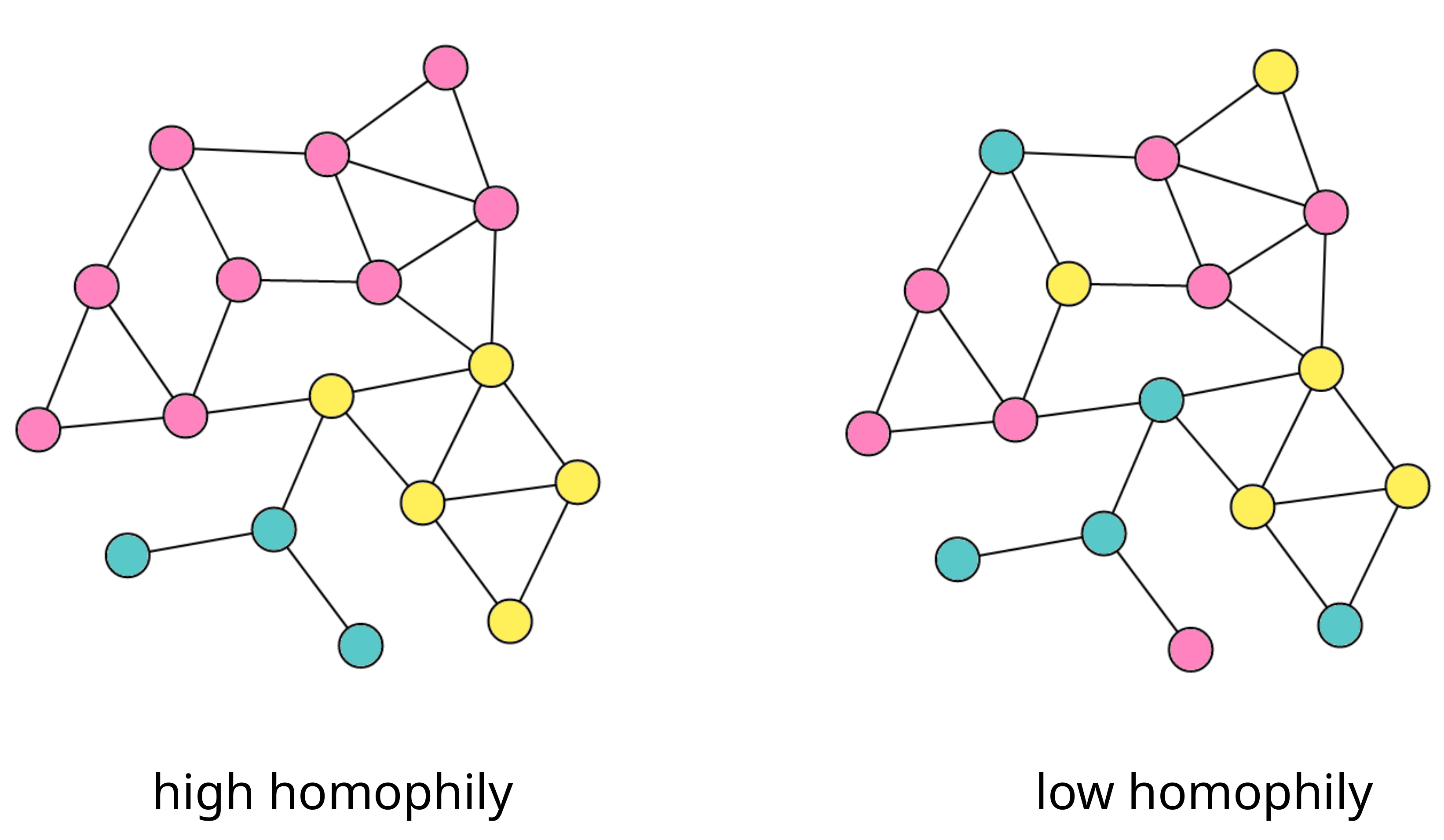}
\caption{Depending on the label assignment $y_v$, the same underlying graph $\mathcal{G}$ can have high homophily ($\mathfrak{h}_\mathcal{G} = 0.85$, left) or low homophily ($\mathfrak{h}_\mathcal{G} = 0.46$, right).}
\label{fig:homophily}
\end{figure}

As the development of deep learning on graphs progressed, several challenges preventing the computation of effective node representations have emerged.
Li et al. \cite{Li2018} have shown that stacking more than three or four layers of graph convolution causes a degradation in accuracy, since representations $\mathbf{h}_v^{(\ell)}$ converge asymptotically to a fixed point of $\mathbf{\hat{A}}$ as $\ell$ increases, or more generally, to a low-frequency subspace of the graph spectrum \cite{Oono2020}.
This problem is known as over-smoothing.
Indeed, by acting as a low-pass filter, GCNs are biased in favor of tasks whose graphs present a high degree of homophily, that is nodes in the same neighborhood mostly share the same class \citep{Zhu2020}.
Formally, homophily in a graph can be quantified as the intra-class edges ratio
\begin{equation}\label{eq:homophily}
	\mathfrak{h}_\mathcal{G} = \frac{\left|\{(v, u) \in \mathcal{E} : y_v = y_u\}\right|}{\left|\mathcal{E}\right|}.
\end{equation}
Alternative homophily measures have also been defined, such as the average of node neighbourhood homophily ratios $\mathfrak{h}_v = |\{ u \in \mathcal{N}_1(v) : y_v = y_u \}| / |\mathcal{N}_1(v)|$ \cite{Pei2020}, or the excess homophily with respect to a random graph connectivity \cite{Lim2021}.
The homophily of a graph is not an intrinsic property of graph connectivity, but it depends on the particular node label assignment $y_v$, as Fig.~\ref{fig:homophily} shows.

Over-squashing instead is an issue that intrinsically depends on message-passing that follows the local graph connectivity: the problem of encoding an exponentially growing receptive field \citep{Micheli2009} in a fixed-size node embedding dimension.
This problem can lead to the inability to extract meaningful features in deeper layers, which along with over-smoothing causes the under-reaching in tasks that require discovering long-range relationships between nodes.
The latter problem is often more relevant in low-homophily settings, since most nodes sharing the same labels are not neighbours.
Topping et al. \cite{Topping2022} have provided theoretical insights into this issue by identifying over-squashing with the exponential decrease in sensitivity of node representations to the input features on distant nodes, as the number of layers increases.
For example, in a GCN model \cite{Kipf2017} the sensitivity of $\mathbf{h}_v^{(L)}$ to the input $\mathbf{x}_{u}$, assuming that there exists an $L$-path between nodes $v$ and $u$, is upper bounded by
\begin{equation}\label{eq:sensitivity}
	\left\lVert \frac{\partial \mathbf{h}_v^{(L)}}{\partial \mathbf{x}_{u}} \right\rVert \leq \left(\prod_{\ell=1}^{L} \lVert \mathbf{W}^{(\ell)} \rVert \right)\, (\mathbf{\hat{A}}^{L})_{v,u}.
\end{equation}
Graph rewiring methods aim to improve this bound by acting on the term $(\mathbf{\hat{A}}^{L})_{v,u}$, for example guided by graph curvature \cite{Topping2022}.

As the reader may have noticed, the challenges to learning effective node representations are of various nature, and all contribute to the difficulty of heterophilic tasks.
In end-to-end trained models, there is also the additional difficulty of back-propagating the gradient through many message-passing layers.
Discerning the inter-play between all these issues is still an open research question.

\paragraph{Proposed solutions}

Changes in the model architectural bias have been proposed to improve classification on heterophilic graphs.
Some solutions identified by \cite{Zhu2020} are:
\begin{enumerate}
	\item Separate ego and neighbourhood representations in \eqref{eq:gnn-layer}, by aggregating on open node neighbourhoods $\mathcal{N}_r(v) \setminus \{v\}$ and combining by concatenation, as done by GraphSAGE \cite{Hamilton2017}.
	This design allows the model to choose how much to take into account neighbouring nodes in computing the node representation.
	\item Extend aggregation to multi-hop neighbourhoods $\mathcal{N}_r(v)$, $r > 1$, such as in graph convolutions that implement Chebyshev polynomial filters (e.g. ChebNet \cite{Defferrard2016}) or that aggregate with powers of the adjacency matrix (e.g. MixHop \cite{AbuElHaija2019}).
	This design aims to improve the local homophily ratio when performing neighbourhood aggregation.
	\item Exploit also the representations $\mathbf{h}_v^{(\ell)}$ computed at each intermediate layer $\ell < L$ to make predictions, e.g. as in Jumping-Knowledge (JK) networks \cite{Xu2018}.
	This choice also partly addresses having an excessive smoothed representation in deeper layers; other models aim to achieve a similar effect via skip-connections instead (e.g. GCNII \cite{ChenMing2020}).
\end{enumerate}
H2GCN \cite{Zhu2020} incorporates all three architectural solutions.
Other models perform multi-hop aggregations via personalized PageRank (APPNP \cite{Klicpera2019}) or a generalized version thereof (GPR-GNN \cite{Chien2021}).
Alternative solutions include altering the graph structure to improve the homophily degree, in order to increase the ratio of intra-class edges in node neighbourhoods \cite{Gasteiger2019}.
Models such as LINK \cite{Zheleva2009} and LINKX \cite{Lim2021} abandon message-passing altogether, by directly using the adjacency matrix rows as input features to an MLP, thus also losing adaptiveness to graph changes and invariance to node ordering.

\section{Reservoir Computing for Graph Nodes}
\label{sec:gesn}

\begin{figure*}
	\centering
	\includegraphics[scale=.2]{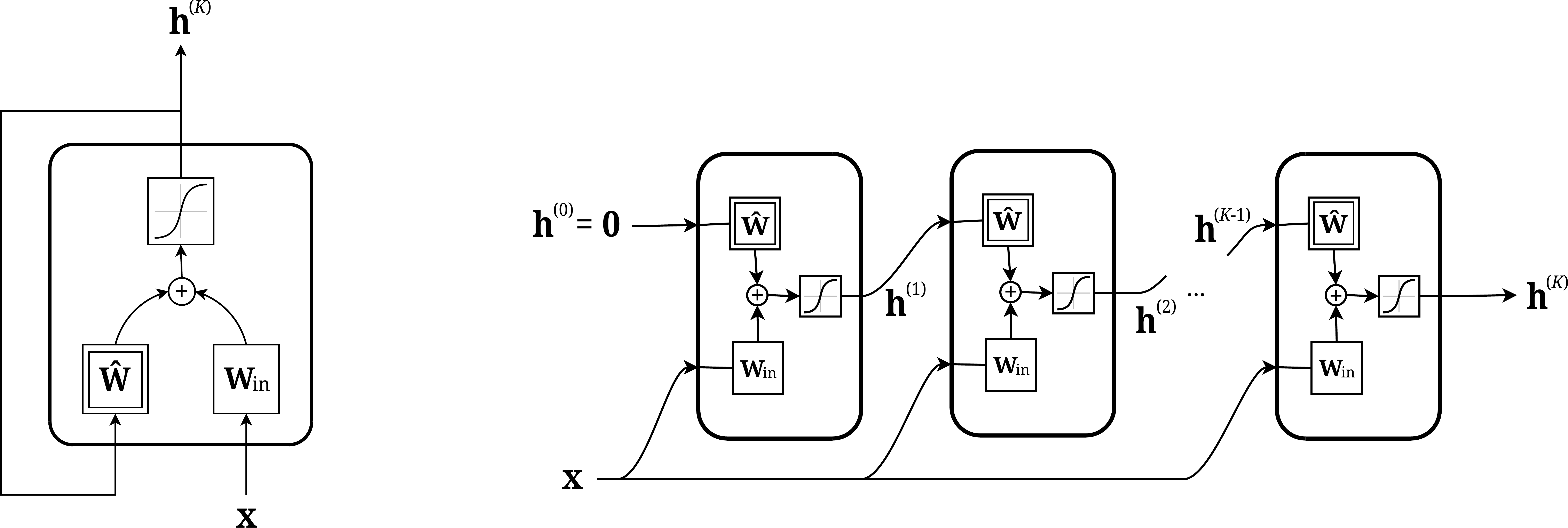}
	\caption{The dynamical system defined in equation \eqref{eq:graphesn} that computes GESN state $\mathbf{h}^{(K)}$ (on the left side). The unfolding of the recursive computation (on the right side) can be interpreted as a deep graph convolutional network made of a random projection (for $k=1$, where no effective contribution is coming from neighbourhood aggregation due to the null state initialization $\mathbf{h}^{(0)} = \mathbf{0}$) followed by $K-1$ convolutional layers with shared weights $\mathbf{W}_{\mathrm{in}}, \mathbf{\hat{W}}$ and input skip-connections. Inside the layer functional blocks, a single stroke rectangle represents the linear function $\mathbf{W}_{\mathrm{in}}$ applied to all node features, while a double-stroke rectangle represents a GCN-like graph convolution with neighbourhood aggregation weights $\mathbf{\hat{W}}$.}
	\label{fig:gesn}
\end{figure*}

Reservoir computing \citep{Nakajima2021,Lukosevicius2009,Verstraeten2007} is a paradigm for the design of efficient recurrent neural networks (RNNs).
Input data is encoded by a randomly initialized reservoir, while only the readout layer for downstream task predictions requires training.
Reservoir computing models, in particular Echo State Networks (ESNs) \citep{Jaeger2004}, have been studied in order to obtain insights into the architectural bias of RNNs \citep{Hammer2003,Gallicchio2011}.

\subsection{Graph Echo State Network for node classification}

GESN has been introduced by Gallicchio and Micheli \cite{Gallicchio2010}, extending the reservoir computing paradigm to graph-structured data.
This model has already demonstrated its effectiveness in graph-level classification tasks \citep{Gallicchio2020}, where a whole-graph embedding is computed from node embeddings via a parameter-free pooling function, such as sum or mean.
Node embeddings are in turn obtained as the state of a non-linear dynamical system, which is iteratively updated in a recursive fashion, similarly to the \textsl{Graph Neural Network} model \cite{Scarselli2009}.
In detail, node embeddings $\mathbf{h}_v^{(k)} \in \mathbb{R}^H$ are recursively computed by the non-linear dynamical system
\begin{equation}\label{eq:graphesn}
	\mathbf{h}_v^{(k)} = \tanh\left(\mathbf{W}_{\mathrm{in}}\, \mathbf{x}_v + \sum_{u \in \mathcal{N}_1(v)} \mathbf{\hat{W}}\, \mathbf{h}_{u}^{(k-1)}\right),\quad \mathbf{h}_v^{(0)} = \mathbf{0},
\end{equation}
where $\mathbf{W}_{\mathrm{in}} \in \mathbb{R}^{H \times X}$ and $\mathbf{\hat{W}} \in \mathbb{R}^{H \times H}$ are the input-to-reservoir and the recurrent weights, respectively, for a reservoir with $H$ units (input bias is omitted).
Reservoir weights are randomly initialized from a uniform distribution in $[-1,1]$, and then rescaled to the desired input scaling and reservoir spectral radius, without requiring any training.
Equation \eqref{eq:graphesn} is iterated over $k$ up to $K$ times, then the final state $\mathbf{h}_v^{(K)}$ is used as the node embedding.
For node classification tasks, a linear readout is applied to node embeddings $\mathbf{y}_v = \mathbf{W}_\mathrm{out}\, \mathbf{h}_v^{(K)} + \mathbf{b}_\mathrm{out}$, where the weights $\mathbf{W}_\mathrm{out} \in \mathbb{R}^{C \times H}, \mathbf{b}_\mathrm{out} \in \mathbb{R}^C$ are trained by ridge regression on one-hot encodings of target classes $y_v$.

In the previous literature, the dynamical system \eqref{eq:graphesn} was constrained to be asymptotically stable, that is to converge to a fixed point $\mathbf{h}_v^{(\infty)}$ as $K \to \infty$.
The existence of a fixed point is guaranteed by the Graph Embedding Stability (GES) property \cite{Gallicchio2020}, which also guarantees independence from the system's initial state $\mathbf{h}_v^{(0)}$.
A sufficient condition for the GES property is requiring that the transition function defined in \eqref{eq:graphesn} to be contractive, i.e. to have Lipschitz constant $\lVert \mathbf{\hat{W}} \rVert \, \lVert \mathbf{A} \rVert < 1$.
In standard reservoir computing practice, however, the recurrent weights are initialized according to a necessary condition \cite{Tortorella2022} for the GES property, which is $\rho(\mathbf{\hat{W}}) < 1 / \alpha$, where $\rho(\cdot)$ denotes the spectral radius of a matrix, i.e. its largest absolute eigenvalue, and $\alpha = \rho(\mathbf{A})$ is the graph spectral radius.
This condition provides the best estimate of the system bifurcation point, i.e. the threshold beyond which \eqref{eq:graphesn} becomes asymptotically unstable \citep{Tortorella2022}.
Previous literature has also shown that on graph-level tasks, the system stability is essential to provide global graph embeddings $\mathbf{h}_\mathcal{G}$ via parameter-free pooling functions, e.g. sum pooling: $\mathbf{h}_\mathcal{G} = \sum_{v \in \mathcal{V}} \mathbf{h}_v^{(\infty)}$.

\subsection{Beyond stability constraints}

Let us now consider a GESN where the number of iterations of \eqref{eq:graphesn} is fixed to a constant $K$.
In this case, the $K$ iterations of the state transition function \eqref{eq:graphesn} can be interpreted as equivalent to $L = K - 1$ graph convolution layers with weights shared among layers and input skip connections.
This interpretation is illustrated in Fig.~\ref{fig:gesn}: the functional blocks represent the unfolding of the recursive state computation in equivalent layers (on the right), where the double border represent a GCN-like graph convolution with neighbourhood aggregation weights $\mathbf{\hat{W}}$.
Notice that at iteration $k = 1$ no convolution is effectively performed, since all node states are null initialized.

For GESN, the sensitivity of $\mathbf{h}_v^{(K)}$ to the input $\mathbf{x}_{u}$ is upper bounded by
\begin{equation}\label{eq:gesn-sensitivity}
	\left\lVert \frac{\partial \mathbf{h}_v^{(K)}}{\partial \mathbf{x}_{u}} \right\rVert \leq \sum_{\ell=0}^{K-1} \lVert \mathbf{\hat{W}} \rVert^{\ell}\, \lVert \mathbf{W}_{\mathrm{in}} \rVert\, (\mathbf{A}^{\ell})_{v,u},
\end{equation}
where by $\lVert \cdot \rVert$ we assume the $2$-norm.
Comparing equation \eqref{eq:gesn-sensitivity} with the input sensitivity in GCN \eqref{eq:sensitivity}, notice the effect of the input term in the state transition function \eqref{eq:graphesn}, which acts as an input skip connection.
While in \eqref{eq:sensitivity} only paths of length $L$ contribute to the node embeddings, in GESN all paths up to length $K-1$ between nodes $u$ and $v$ are taken into account.
However, not all the paths influence equally $\mathbf{h}_v^{(K)}$.
According to the value of $\lVert \mathbf{\hat{W}} \rVert \, \lVert \mathbf{A} \rVert$, we have two cases:
\begin{enumerate}
	\item If $\lVert \mathbf{\hat{W}} \rVert \, \lVert \mathbf{A} \rVert < 1$, then the contribution from longer paths is exponentially vanishing, as $\lVert \mathbf{\hat{W}} \rVert^\ell \, \lVert \mathbf{A} \rVert^\ell \to 0$ for $\ell \to \infty$. This is the case where GESN satisfies the GES property, as its dynamics are contractive and the embeddings converge to a fixed point $\mathbf{h}_v^{(\infty)}$.
	\item Otherwise, if $\lVert \mathbf{\hat{W}} \rVert \, \lVert \mathbf{A} \rVert \geq 1$, the GES property is no longer guaranteed to be satisfied, and the dynamics may present orbits or chaotic behaviour for $\lVert \mathbf{\hat{W}} \rVert \, \lVert \mathbf{A} \rVert \gg 1$. However, equation \eqref{eq:gesn-sensitivity} suggests that an initialization which violates the sufficient condition for the GES property is necessary to prevent the contributions from longer paths to be exponentially vanishing.
\end{enumerate}
In our network, we are able to explicitly choose how large the layers' Lipschitz constant $\lVert \mathbf{\hat{W}} \rVert$ is, thus controlling both the dynamical and the sensitivity behaviour.
Evaluating the effectiveness of node embeddings obtained from non-contractive dynamics is one of the aims of our work, which will be carried on experimentally in section~\ref{sec:experiments} on several high and low homophily node classification tasks.
To investigate the different dynamics of GESN, we follow the standard practice in reservoir computing of selecting the spectral radius $\rho(\mathbf{\hat{W}})$ of recurrent weights initializations.
The spectral radius is a lower bound for the spectral norm \cite{Goldberg1974}, i.e. $\lVert \mathbf{\hat{W}} \rVert \geq \rho(\mathbf{\hat{W}})$, and can be set very efficiently \cite{Gallicchio2020inns}.
Since $\lVert \mathbf{\hat{W}} \rVert \, \lVert \mathbf{A} \rVert \geq \rho(\mathbf{\hat{W}})\, \alpha$, non-contractive dynamics will be found in the region $\rho(\mathbf{\hat{W}}) > 1 / \alpha$.

Our approach is alternative to graph rewiring methods, which act directly on the term dependent on the graph adjacency $\mathbf{A}$ instead.
As argued by \cite{Topping2022}, the exponentially vanishing sensitivity in e.g. equation \eqref{eq:sensitivity} could be caused by topological bottlenecks in the factor $(\mathbf{\hat{A}}^{L})_{v,u}$.
We defer a discussion of rewiring methods to section~\ref{sec:experiments-rewiring}, where we will also experimentally investigate the hypothesis that this problem could be contrasted by going beyond a contractive initialization via $\lVert \mathbf{\hat{W}} \rVert \, \lVert \mathbf{A} \rVert > 1$ as we have just proposed.

\section{Experiments}
\label{sec:experiments}

In this section we evaluate GESN on three sets of node classification tasks with varying degrees of homophily and ranging from small to large graphs, as summarised in Tab.~\ref{tab:task-stats}, for a total of 19 tasks extracted from 13 real-world graphs.
We compare the classification accuracy with a variety of fully-trained deep models, which were presented in section \ref{sec:background}.
In all tasks, we select by grid search the reservoir radius, input scaling factor, number of hidden units, and readout regularization in ridge regression, while keeping the number of iterations of equation \eqref{eq:graphesn} fixed at $K = 100$.
As our analysis in section \ref{sec:analysis} will demonstrate, the number of iterations is not much relevant, provided that $K$ is large enough for GESN to capture the whole graph structure.
Code for reproducing our experiments is publicly available.\footnote{\url{https://github.com/dtortorella/addressing-heterophily-gesn}}

\begin{table*}
	\centering
	\begin{tabular}{lcrrrrc}
		\toprule
		\textbf{Graph} & \textbf{Homophily} & \textbf{Nodes} & \textbf{Edges} & \textbf{Radius} $\alpha$ & \textbf{Features} & \textbf{Classes} \\
		\midrule
		Texas & $0.11$ & $183$ & $295$ & $2.56$ & $1\mathord{,}703$ & $5$ \\
		Wisconsin & $0.21$ & $251$ & $466$ & $2.88$ & $1\mathord{,}703$ & $5$ \\
		Actor & $0.22$ & $7\mathord{,}600$ & $26\mathord{,}752$ & $9.99$ & $932$ & $5$ \\
		Squirrel & $0.22$ & $5\mathord{,}201$ & $198\mathord{,}493$ & $138.60$ & $2\mathord{,}089$ & $5$ \\
		Chameleon & $0.23$ & $2\mathord{,}277$ & $31\mathord{,}421$ & $61.90$ & $2\mathord{,}089$ & $5$ \\
		Cornell & $0.30$ & $183$ & $280$ & $2.68$ & $1\mathord{,}703$ & $5$ \\
		\midrule
		Citeseer & $0.74$ & $3\mathord{,}327$ & $9\mathord{,}104$ & $13.74$ & $3\mathord{,}703$ & $6$ \\
		Pubmed & $0.80$ & $19\mathord{,}717$ & $88\mathord{,}648$ & $23.24$ & $500$ & $3$ \\
		Cora & $0.81$ & $2\mathord{,}708$ & $10\mathord{,}556$ & $14.39$ & $1\mathord{,}433$ & $7$ \\
		\midrule
		Penn94 & $0.47$ & $41\mathord{,}554$ & $1\mathord{,}362\mathord{,}229$ & $180.44$ & $4\mathord{,}814$ & $2$ \\
		arXiv-year & $0.22$ & $169\mathord{,}343$ & $1\mathord{,}166\mathord{,}243$ & $8.96$ & $128$ & $5$ \\
		genius & $0.62$ & $421\mathord{,}961$ & $984\mathord{,}979$ & $212.82$ & $12$ & $2$ \\
		twitch-gamers & $0.55$ & $168\mathord{,}114$ & $6\mathord{,}797\mathord{,}557$ & $149.92$ & $2\mathord{,}514$ & $2$ \\
		\bottomrule
	\end{tabular}
\caption{Statistics on the graphs employed in the node classification tasks.}
\label{tab:task-stats}
\end{table*}

\begin{figure}
	\centering
	\includegraphics[scale=.8,trim=0cm 1.5cm 0 0,clip]{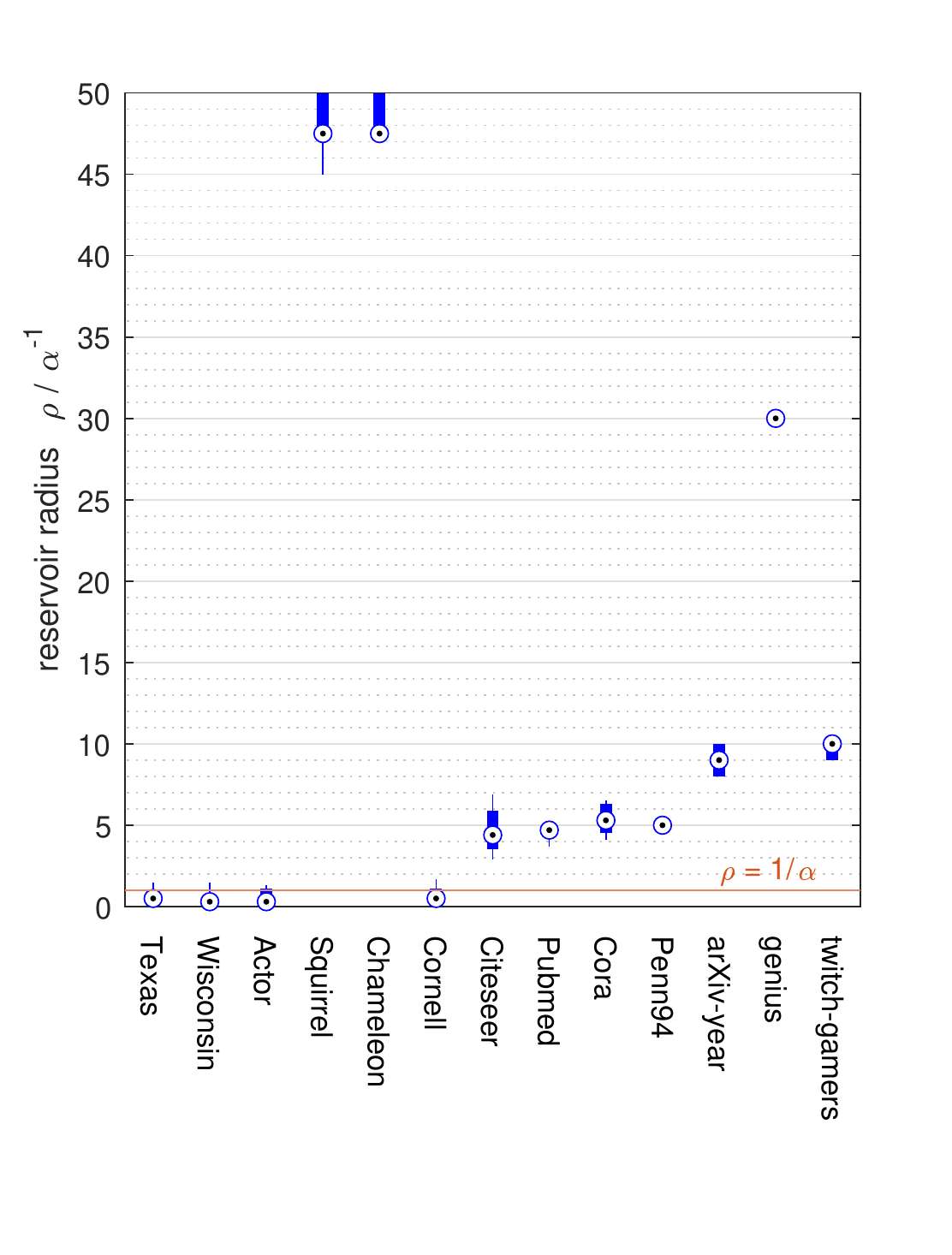}
	\caption{Normalized spectral radii $\rho(\mathbf{\hat{W}})$ selected in the node classification experiments via hyper-parameter grid search on GESN.}
	\label{fig:radii}
\end{figure}

\subsection{Medium-scale tasks}
\label{sec:experiments-medium}


\begin{table*}
	\centering
	\begingroup
	\setlength{\tabcolsep}{4pt}
	\begin{tabular}{lccccccccc}
		\toprule
		& \textbf{Texas} & \textbf{Wisconsin} & \textbf{Actor} & \textbf{Squirrel} & \textbf{Chameleon} & \textbf{Cornell} & \textbf{Citeseer} & \textbf{Pubmed} & \textbf{Cora} \\
		\midrule
		GCN & $59.46_{\pm 5.25}$ & $59.80_{\pm 6.99}$ & $30.26_{\pm 0.79}$ & $36.89_{\pm 1.34}$ & $59.82_{\pm 2.58}$ & $57.03_{\pm 4.67}$ & $\mathbf{76.68}_{\pm 1.64}$ & $87.38_{\pm 0.66}$ & $\mathbf{87.28}_{\pm 1.26}$ \\
		GAT & $58.38_{\pm 4.45}$ & $55.29_{\pm 8.71}$ & $26.28_{\pm 1.73}$ & $30.62_{\pm 2.11}$ & $54.69_{\pm 1.95}$ & $58.92_{\pm 3.32}$ & $\mathbf{75.46}_{\pm 1.72}$ & $84.68_{\pm 0.44}$ & $82.68_{\pm 1.80}$ \\
		GraphSAGE & $\mathbf{82.43}_{\pm 6.14}$ & $81.18_{\pm 5.56}$ & $34.23_{\pm 0.99}$ & $41.61_{\pm 0.74}$ & $58.73_{\pm 1.68}$ & $75.95_{\pm 5.01}$ & $\mathbf{76.04}_{\pm 1.30}$ & $88.45_{\pm 0.50}$ & $\mathbf{86.90}_{\pm 1.04}$ \\
		\midrule
		GCN+JK & $66.49_{\pm 6.64}$ & $74.31_{\pm 6.43}$ & $34.18_{\pm 0.85}$ & $40.45_{\pm 1.61}$ & $63.42_{\pm 2.00}$ & $64.59_{\pm 8.68}$ & $74.51_{\pm 1.75}$ & $88.41_{\pm 0.45}$ & $85.79_{\pm 0.92}$ \\
		GCN+Cheby & $77.30_{\pm 4.07}$ & $79.41_{\pm 4.46}$ & $34.11_{\pm 1.09}$ & $43.86_{\pm 1.64}$ & $55.24_{\pm 2.76}$ & $74.32_{\pm 7.46}$ & $\mathbf{75.82}_{\pm 1.53}$ & $88.72_{\pm 0.55}$ & $\mathbf{86.76}_{\pm 0.95}$ \\
		GraphSAGE+JK & $\mathbf{83.78}_{\pm 2.21}$ & $81.96_{\pm 4.96}$ & $34.28_{\pm 1.01}$ & $40.85_{\pm 1.29}$ & $58.11_{\pm 1.97}$ & $75.68_{\pm 4.03}$ & $\mathbf{76.05}_{\pm 1.37}$ & $88.34_{\pm 0.62}$ & $85.96_{\pm 0.83}$ \\
		\midrule
		MixHop & $77.84_{\pm 7.73}$ & $75.88_{\pm 4.90}$ & $32.22_{\pm 2.34}$ & $43.80_{\pm 1.48}$ & $60.50_{\pm 2.53}$ & $73.51_{\pm 6.34}$ & $\mathbf{76.36}_{\pm 1.33}$ & $85.31_{\pm 0.61}$ & $\mathbf{87.61}_{\pm 0.85}$ \\
		H2GCN & $\mathbf{84.86}_{\pm 6.77}$ & $\mathbf{86.67}_{\pm 4.69}$ & $\mathbf{35.86}_{\pm 1.03}$ & $36.42_{\pm 1.89}$ & $57.11_{\pm 1.58}$ & $\mathbf{82.16}_{\pm 4.80}$ & $\mathbf{77.07}_{\pm 1.64}$ & $\mathbf{89.40}_{\pm 0.34}$ & $\mathbf{86.92}_{\pm 1.37}$ \\
		LINKX & $74.60_{\pm 8.37}$ & $75.49_{\pm 5.72}$ & $\mathbf{36.10}_{\pm 1.55}$ & $61.81_{\pm 1.80}$ & $68.42_{\pm 1.38}$ & $\mathbf{77.84}_{\pm 5.81}$ & $73.19_{\pm 0.99}$ & $87.86_{\pm 0.77}$ & $84.64_{\pm 1.13}$ \\
		\midrule
		MLP & $\textbf{81.89}_{\pm 4.78}$ & $\mathbf{85.29}_{\pm 3.61}$ & $\mathbf{35.76}_{\pm 0.98}$ & $29.68_{\pm 1.81}$ & $46.36_{\pm 2.52}$ & $\mathbf{81.08}_{\pm 6.37}$ & $72.41_{\pm 2.18}$ & $86.65_{\pm 0.35}$ & $74.75_{\pm 2.22}$ \\
		\midrule
		GESN & $\textbf{84.31}_{\pm 4.44}$ & $\mathbf{83.33}_{\pm 3.81}$ & $\mathbf{34.56}_{\pm 0.76}$ & $\mathbf{73.56}_{\pm 1.62}$ & $\mathbf{77.05}_{\pm 1.24}$ & $\mathbf{81.14}_{\pm 6.00}$ & $74.51_{\pm 2.14}$ & $\mathbf{89.20}_{\pm 0.34}$ & $86.04_{\pm 1.01}$ \\
		\bottomrule
	\end{tabular}
	\endgroup
\caption{Node classification accuracy on low and high homophily graphs following the experimental setting of \cite{Zhu2020}. Average accuracy and standard deviation for GESN is reported from \cite{Tortorella2022esann}, while LINKX is reported from \cite{Lim2021} and other models are reported from \cite{Zhu2020}. Results within one standard deviation of the best accuracy are highlighted.}
\label{tab:experiments-esann}
\end{table*}

\begin{figure*}
	\centering
	\includegraphics[scale=.75,trim=2cm .5cm 0 .75cm,clip]{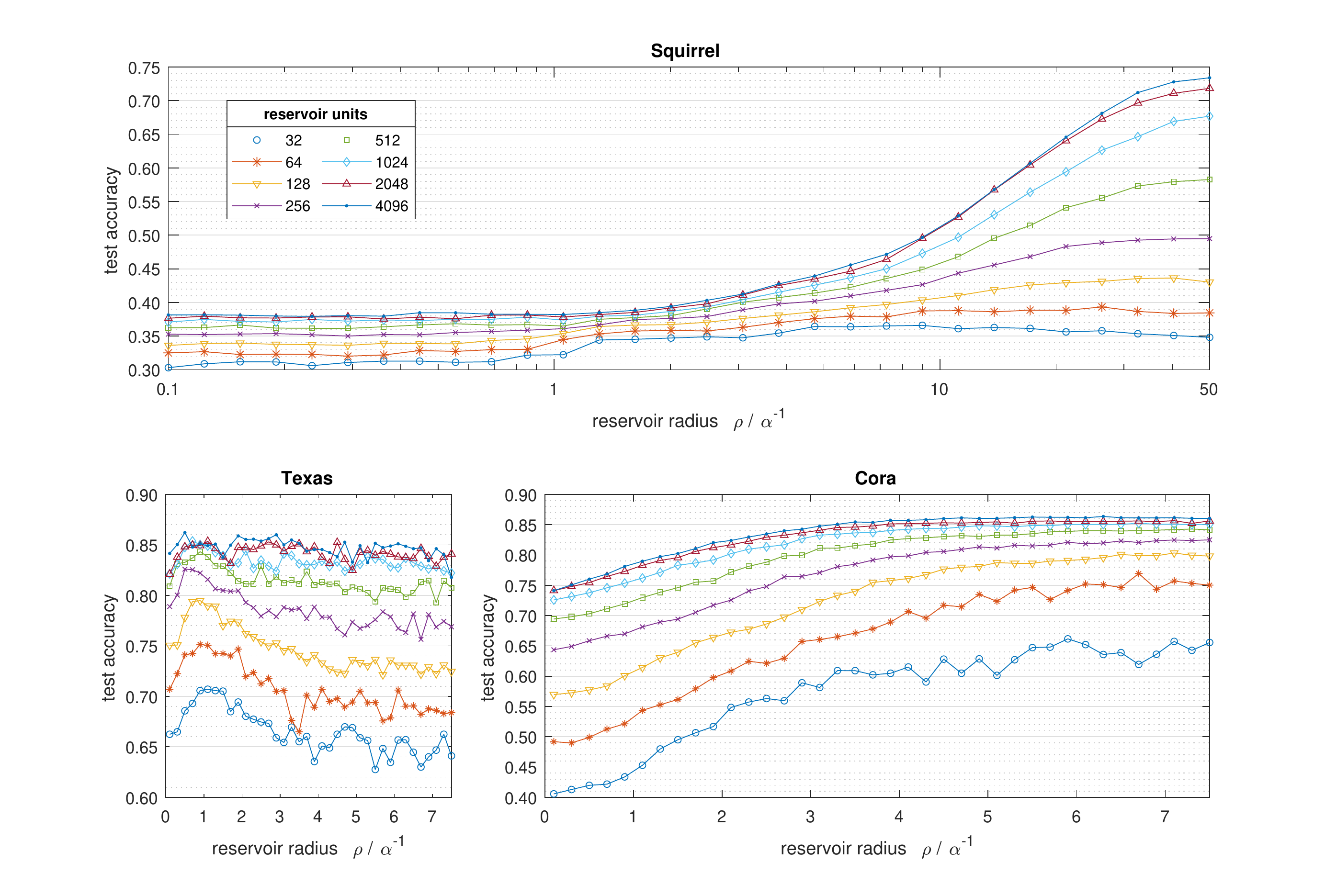}
	\caption{Impact of reservoir radius and number of reservoir units in GESN on three medium-scale tasks. (Best viewed in colour.)}
	\label{fig:esann-experiments}
\end{figure*}

\begin{figure*}
	\centering
	\includegraphics[scale=.75,trim=2cm .5cm 0 .5cm,clip]{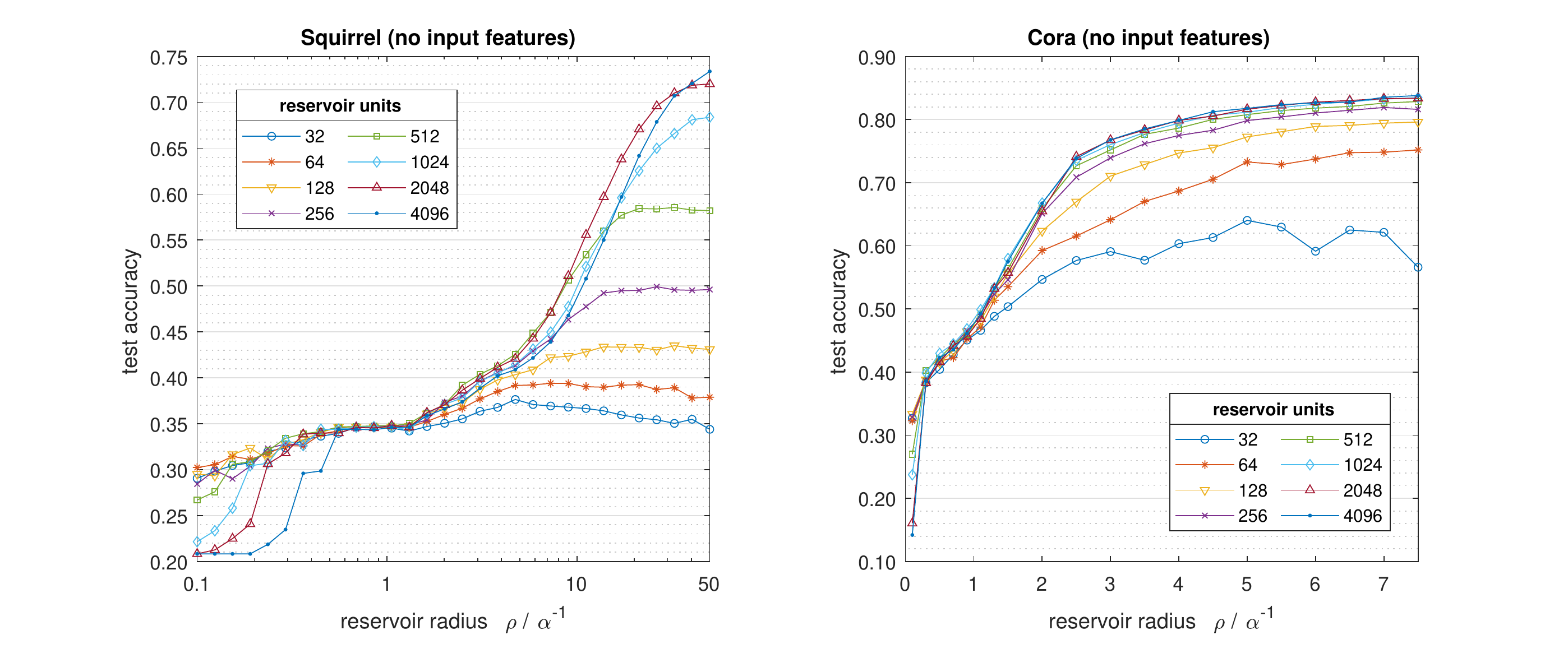}
	\caption{Impact of reservoir radius and number of reservoir units in GESN on two medium-scale tasks where original input features have been replaced with a constant value on all nodes. (Best viewed in colour.)}
	\label{fig:esann-no-features}
\end{figure*}

We begin our evaluation from nine small- to medium-scale (up to $10^3$ nodes) classification tasks which are widely adopted in literature.
Namely, we evaluate GESN on six node classification tasks with low homophily degree ($\mathfrak{h}_\mathcal{G} \leq 0.3$) and three tasks with high homophily degree ($\mathfrak{h}_\mathcal{G} > 0.7$).
We adopt the same $10$ scaffold splits 48\%/32\%/20\% of \cite{Zhu2020}, averaging results in each fold over $10$ different reservoir initializations.
We explore a number of units ranging from $2^4$ to $2^{12}$, input scaling factors from $1$ to $\frac{1}{320}$, readout regularization values from $10^{-5}$ to $10^2$, and reservoir radii $\rho$ from $0.1/\alpha$ to $50/\alpha$ (obtained via grid search).

Accuracy results are reported in Tab.~\ref{tab:experiments-esann}, while Fig.~\ref{fig:radii} shows the reservoir radii selected in the $10$ splits.
We can observe three different behaviours, exemplified in Fig.~\ref{fig:esann-experiments}.
The number of reservoir units plays a significant role, offering best results when it is closer to the number of input features.
For Texas, Wisconsin, Actor, and Cornell, the performances of GESN are closer to the accuracies of MLP, which uses only node features $\mathbf{x}_v$, and H2GCN, with reservoir radii $\rho < 1/\alpha$: in this case, the graph connectivity appears to be of no use.
Other convolutive models, even with the addition of some architectural variation, perform significantly worse in general.
While Squirrel and Chameleon present a low homophily degree, graph convolution models fare better than MLP: in this case graph connectivity needs to be taken into account.
On these two tasks, GESN improves upon the best message-passing model accuracy respectively by $27.3\%$ and $12.8\%$ (by $14.8\%$ and $8.7\%$ with respect to LINKX, which is not adaptive), with much larger reservoir radii selected in the range $45/\alpha$ to $50/\alpha$.
Finally, on high homophily tasks (Citeseer, Pubmed, Cora) GESN performs generally in line with graph convolution models, which in turn do better than MLP; reservoir radii are selected in the range $4/\alpha$ to $6/\alpha$.

We observe how the best accuracy results are for reservoir radii well above the stability threshold, which appear to be required when the graph connectivity needs to be leveraged in classifying nodes.
To support this conclusion, we analyse how replacing the original node input features with constant ones affects the accuracy, since in this case GESN can rely only to graph connectivity for computing distinguishable node embeddings.
In Fig.~\ref{fig:esann-no-features} we report the accuracy on Squirrel and Cora where input features have thus been removed.
We observe that for stable embeddings ($\rho < 1 / \alpha$), accuracy significantly drops below the level reached by having input features, while it reaches almost the same levels of accuracy for the values of $\rho$ selected in tasks with original node features, which are well beyond the region where GESN stability is guaranteed.
As we pointed out in section \ref{sec:gesn} concerning equation \eqref{eq:gesn-sensitivity}, a non-contractive reservoir initialization (such as $\rho(\mathbf{\hat{W}}) > 1/\alpha$) prevents the contribution from longer paths to be exponentially diminishing, thus enabling GESN to capture structural relationships of a node with respect to a larger sub-graph.

\subsection{Large-scale tasks}
\label{sec:experiments-large}


\begin{table*}
	\centering
	\begin{tabular}{lcccc}
		\toprule
		 & \textbf{Penn94} & \textbf{arXiv-year} & \textbf{genius} & \textbf{twitch-gamers} \\
		\midrule
		MLP & $73.61_{\pm 0.40}$ & $36.70_{\pm 0.21}$ & $86.68_{\pm 0.09}$ & $60.92_{\pm 0.07}$ \\
		\midrule
		L. Prop. 1-hop & $63.21_{\pm 0.39}$ & $43.42_{\pm 0.17}$ & $66.02_{\pm 0.16}$ & $62.77_{\pm 0.24}$ \\
		L. Prop. 2-hop & $74.13_{\pm 0.46}$ & $46.07_{\pm 0.15}$ & $67.04_{\pm 0.20}$ & $63.88_{\pm 0.24}$ \\
		SGC 1-hop & $66.79_{\pm 0.27}$ & $32.83_{\pm 0.13}$ & $82.36_{\pm 0.37}$ & $58.97_{\pm 0.19}$ \\
		SGC 2-hop & $76.09_{\pm 0.45}$ & $32.27_{\pm 0.06}$ & $82.10_{\pm 0.14}$ & $59.94_{\pm 0.21}$ \\
		C\&S 1-hop & $74.28_{\pm 1.19}$ & $44.51_{\pm 0.16}$ & $82.93_{\pm 0.15}$ & $64.86_{\pm 0.27}$ \\
		C\&S 2-hop & $78.40_{\pm 3.12}$ & $49.78_{\pm 0.26}$ & $84.94_{\pm 0.49}$ & $65.02_{\pm 0.16}$ \\
		\midrule
		GCN & $82.47_{\pm 0.27}$ & $46.02_{\pm 0.26}$ & $87.42_{\pm 0.37}$ & $62.18_{\pm 0.26}$ \\
		GAT & $81.53_{\pm 0.55}$ & $46.05_{\pm 0.51}$ & $55.80_{\pm 0.87}$ & $59.89_{\pm 4.12}$ \\
		GCN+JK & $81.63_{\pm 0.54}$ & $46.28_{\pm 0.29}$ & $89.30_{\pm 0.19}$ & $63.45_{\pm 0.22}$ \\
		GAT+JK & $80.69_{\pm 0.36}$ & $45.80_{\pm 0.72}$ & $56.70_{\pm 2.07}$ & $59.98_{\pm 2.87}$ \\
		\midrule
		H2GCN & OOM & $49.09_{\pm 0.10}$ & OOM & OOM \\
		MixHop & $83.47_{\pm 0.71}$ & $51.81_{\pm 0.17}$ & $90.58_{\pm 0.16}$ & $65.64_{\pm 0.27}$ \\
		APPNP & $74.33_{\pm 0.38}$ & $38.15_{\pm 0.26}$ & $85.36_{\pm 0.62}$ & $60.97_{\pm 0.10}$ \\
		GPR-GNN & $81.38_{\pm 0.16}$ & $45.07_{\pm 0.21}$ & $90.05_{\pm 0.31}$ & $61.89_{\pm 0.29}$ \\
		GCNII & $82.92_{\pm 0.59}$ & $47.21_{\pm 0.28}$ & $90.24_{\pm 0.09}$ & $63.39_{\pm 0.61}$ \\
		\midrule
		LINK & $80.79_{\pm 0.49}$ & $53.97_{\pm 0.18}$ & $73.56_{\pm 0.14}$ & $64.85_{\pm 0.21}$ \\
		LINKX & $\mathbf{84.71}_{\pm 0.52}$ & $\mathbf{56.00}_{\pm 1.34}$ & $90.77_{\pm 0.27}$ & $66.06_{\pm 0.19}$ \\
		\midrule
		GESN & $80.29_{\pm 0.41}$ & $48.80_{\pm 0.22}$ & $\mathbf{91.72}_{\pm 0.08}$ & $\mathbf{68.34}_{\pm 0.86}$ \\
		\bottomrule
	\end{tabular}
\caption{Average test accuracy (area under the ROC curve for genius) and standard deviation on four large-scale low-homophily tasks (best results highlighted). Except for GESN, the other results are reported from \cite{Lim2021}.}
\label{tab:experiments-large}
\end{table*}

We now evaluate the accuracy of GESN on four large-scale (from $10^4$ to $10^5$ nodes) heterophilic node classification tasks proposed by \cite{Lim2021}.
Penn94, genius, and twitch-gamers are binary classification tasks extracted from social networks, where the target label is user gender, active account, and explicit content, respectively; arXiv-year is a citation network where the publication year is to be predicted.
We adopt the same $5$ scaffold splits 50\%/25\%/25\% of \cite{Lim2021}, averaging results in each fold over $10$ different reservoir initializations.
We explore a number of units ranging from $2^4$ to $2^{13}$, input scaling factors from $1$ to $0.01$, readout regularization values from $10^{-5}$ to $10^2$, and reservoir radii $\rho \in [1/\alpha, 30/\alpha]$ (obtained via grid search).

Accuracy results are reported in Tab.~\ref{tab:experiments-large}, while Fig.~\ref{fig:radii} shows the reservoir radii selected in the $5$ splits.
We notice that generally an MLP under-performs convolutional models, thus suggesting that on this set of tasks connectivity is also relevant.
Indeed, in case of GESN, we can observe that reservoir radii are selected well beyond the stability region, analogously to what happened for Squirrel and Chameleon.
GESN achieves top accuracy on genious and twitch-gamers, while on Penn94 and arXiv-year it performs generally in line with fully-trained convolutional models.
The less optimal performance on Penn94 can be possibly attributed to the high-dimensional one-hot node features, which pose a difficulty for the input random projection via $\mathbf{W}_{\mathrm{in}}$, while arXiv-year has paper text embeddings as node input features that GESN could be not properly exploiting for the same reason.
We notice also that H2GCN, which in the previous set of tasks demonstrated to generally outperform other fully-trained models, on large-scale graphs is unable to be trained due to too demanding memory requirements; this could be explained by the architectural choices of the model, which implements all the architectural solutions outlined in section \ref{sec:background}, thus increasing the gradient computation cost.
Label-propagation based models \cite{Zhou2004}, including SGC \cite{Wu2019} and C\&S \cite{Huang2021}, generally do not over-perform an MLP, due to their innate bias towards homophilic graphs.
However, an increase to 2-hop neighbourhoods seems to mitigate such bias, as pointed out in section \ref{sec:background}.
A similar strategy of multi-hop neighbourhood aggregation is adopted by APPNP \cite{Klicpera2019}, which performs message-passing on the propagation matrix computed by a personalised PageRank (PPR) algorithm, and by GPR-GNN \cite{Chien2021}, which adopts a generalised PageRank algorithm instead.
Other fully-trained graph convolutional models seems to perform roughly in the same range, except GAT \cite{Velickovic2018} which present some difficulties on genius and twitch-gamers.
Architectural variations, e.g. JK \cite{Xu2018}, or skip connections featured in GCNII \cite{ChenMing2020}, seem to offer just a slight improvement with respect to basic models such as GCN.
Finally, LINKX \cite{Lim2021} performs as the best (Penn94, arXiv-year) or second-best behind GESN (genious and twitch-gamers).
This model also over-performs its variant LINK \cite{Zheleva2009}, which does not exploit node input features, thus demonstrating their significance to the tasks.
However, we point out that both models are not adaptive to alterations in the input graph, nor they are permutation-invariant with respect to node ordering, since both use an MLP on the adjacency matrix rows as (additional) node input features.
In doing so, it is also lost the locality advantage of convolutional models, which perform aggregations limitedly to a node's neighbourhood, and the model sparsity, since the MLP weight dimensions should scale with the number of graph nodes $|V|$.

\subsection{Comparison with rewiring methods}
\label{sec:experiments-rewiring}

Topping et al. \cite{Topping2022} have further investigated the connection of over-squashing --- as measured by the Jacobian of node representations in \eqref{eq:sensitivity} --- with the graph topology via the term $(\mathbf{\hat{A}}^{L})_{v,u}$, and have identified in negative local graph curvature the cause of `bottlenecks' in message propagation.
In order to remove these bottlenecks, they have proposed rewiring the input graph, i.e. altering the original set of edges as a preprocessing step, via \textsl{Stochastic Discrete Ricci Flow} (SDRF).
This method works by iteratively adding an edge to support the most negatively-curved edge while removing the most positively-curved one according to the \emph{balanced Forman curvature} \citep{Topping2022}, until convergence or a maximum number of iterations is reached.
This rewiring approach can be contrasted to e.g. \textsl{Graph Diffusion Convolution} (DIGL) \cite{Gasteiger2019}, which aims to address the problem of noisy edges in the input graph by altering the connectivity according to a generalized graph diffusion process, such as personalized PageRank (PPR).
Since DIGL has a smoothing effect on the graph adjacency by promoting connectivity between nodes that are a short diffusion distance, it may be more suitable for tasks that present a high degree of homophily \citep{Topping2022}, i.e. graphs with an high ratio of intra-class edges.

\begin{table*}
	\centering
	\begin{tabular}{lcccccc}
		\toprule
		& \textbf{Cornell} & \textbf{Texas} & \textbf{Wisconsin} & \textbf{Chameleon} & \textbf{Squirrel} & \textbf{Actor} \\
		\midrule
		None & $52.69_{\pm 0.21}$ & $61.19_{\pm 0.49}$ & $54.60_{\pm 0.86}$ & $41.80_{\pm 0.41}$ & $39.83_{\pm 0.14}$ & $28.70_{\pm 0.09}$ \\
		Undirected & $53.20_{\pm 0.53}$ & $63.38_{\pm 0.87}$ & $51.37_{\pm 1.15}$ & $42.63_{\pm 0.30}$ & $40.77_{\pm 0.16}$ & $28.10_{\pm 0.11}$ \\
		Fully Adjacent & $58.29_{\pm 0.49}$ & $64.82_{\pm 0.29}$ & $55.48_{\pm 0.62}$ & $42.33_{\pm 0.17}$ & $40.74_{\pm 0.13}$ & $28.68_{\pm 0.16}$ \\
		\midrule
		DIGL (PPR) & $58.26_{\pm 0.50}$ & $62.03_{\pm 0.43}$ & $49.53_{\pm 0.27}$ & $42.02_{\pm 0.13}$ & $34.38_{\pm 0.11}$ & $30.79_{\pm 0.10}$ \\
		DIGL + Undir. & $59.54_{\pm 0.64}$ & $63.54_{\pm 0.38}$ & $52.23_{\pm 0.54}$ & $42.68_{\pm 0.12}$ & $33.36_{\pm 0.21}$ & $29.71_{\pm 0.11}$ \\
		\midrule
		SDRF & $54.60_{\pm 0.39}$ & $64.46_{\pm 0.38}$ & $55.51_{\pm 0.27}$ & $43.75_{\pm 0.31}$ & $40.97_{\pm 0.14}$ & $29.70_{\pm 0.13}$ \\
		SDRF + Undir. & $57.54_{\pm 0.34}$ & $70.35_{\pm 0.60}$ & $61.55_{\pm 0.86}$ & $44.46_{\pm 0.17}$ & $41.47_{\pm 0.21}$ & $29.85_{\pm 0.07}$ \\
		\midrule
		GESN & $\mathbf{69.75}_{\pm 1.11}$ & $\mathbf{73.96}_{\pm 1.45}$ & $\mathbf{77.76}_{\pm 1.68}$ & $\mathbf{50.19}_{\pm 0.65}$ & $\mathbf{42.70}_{\pm 0.29}$ & $\mathbf{35.07}_{\pm 0.24}$ \\
		\bottomrule
	\end{tabular}
\caption{Average test accuracy with $95\%$ confidence intervals (best results in bold). Except for GESN, the other results are reported from \cite{Topping2022}. Tasks are limited to the largest connected component of the original graphs.}
\label{tab:experiments-rewiring}
\end{table*}

We compare the accuracy of GESNs on six low-homophily node classification tasks against different rewiring mechanisms applied in conjunction with fully-trained GCNs.
As pointed out in \cite{Topping2022}, avoiding over-squashing in order to capture long-range dependencies is often more relevant in low-homophily settings, since most nodes sharing the same labels are not neighbours.
In our experiments we follow the same setting and training/validation/test splits of \cite{Gasteiger2019,Topping2022}, with tasks limited to the largest connected component of the original graphs, and report the average accuracy with $95\%$ confidence intervals on $1000$ test bootstraps.
As in previous tasks of section \ref{sec:experiments-medium}, the hyper-parameters selected on the validation splits for GESN are: the reservoir radius $\rho(\mathbf{\hat{W}})$ in the range $[0.1/\alpha,50/\alpha]$, which controls how large the Lipschitz constant of \eqref{eq:graphesn} should be; the input scaling factor of $\mathbf{W}_{\mathrm{in}}$ in the range $[\frac{1}{320},1]$; the number of units $H$ in the range $[2^4,2^{12}]$; and the readout regularization for the ridge regression.

The results are reported in Table~\ref{tab:experiments-rewiring}.
For details on the fully-trained models used in conjunction with rewiring methods and their hyper-parameters, we refer to \cite{Topping2022}, where experimental results are taken from.
We observe that GESNs beat the other models by a significant margin on all the six tasks.
Indeed, DIGL and SDRF offer improvements over the baseline GCN of a few accuracy points on average, usually requiring also that the graph to be made undirected.
In contrast, GESN improves by up to $16\%$ over the best rewiring methods, and by $4$-$6$ points on average.
Notice also that rewiring algorithms, in particular SDRF, can be extremely costly and need careful tuning in model selection, in contrast to the efficiency of the reservoir computing approach, which ditches both the preprocessing of input graphs and the training of the node embedding function.
Indeed, just the preprocessing step of SDRF can require computations ranging from the order of minutes to hours, while a complete model can be obtained with GESN in a few seconds' time on the same GPU.

\subsection{Efficiency}

Finally, we evaluate the efficiency of GESN.
Our model is constituted by an untrained module that computes node embeddings recursively via $K$ message-passing operations, and a trained linear readout module.
As for any message-passing model, the cost of a graph convolution is $O(|\mathcal{E}|)$ vector sums.
Since input projection can be computed just once before iterating convolutions, the overall computational complexity of equation \eqref{eq:graphesn} is $O\left(H X |\mathcal{V}| + K (H |\mathcal{V}| + |\mathcal{E}| H + H^2 |\mathcal{V}|)\right)$, which is similar to other convolutional-based models (where $K$ is replaced by the number of layers $L$).
The cost $O(K H^2 |\mathcal{V}|)$ due to state projection can be further reduced to linear with respect to embedding dimension $H$ by introducing sparsity in the recurrent matrix connectivity, as is common practice in reservoir computing \cite{Gallicchio2020}.
In GESN, only the linear readout's $C (H+1)$ parameters require training, against the additional $O(H^2 L)$ parameters of models that need to be trained end-to-end through many gradient descent epochs.
Training the readout via ridge regression can be done efficiently even for large data in a single pass \cite{Zhang2017}, requiring $O\left((H^2 + C H + C^2) |\mathcal{V}|\right)$ to compute a sufficient statistics matrix and at most $O(H^3)$ to solve $C$ linear systems with a direct method (iterative solvers can further lower the computational cost).

\begin{figure}
	\centering
	\includegraphics[scale=.7,trim=0.9cm 0 0 0,clip]{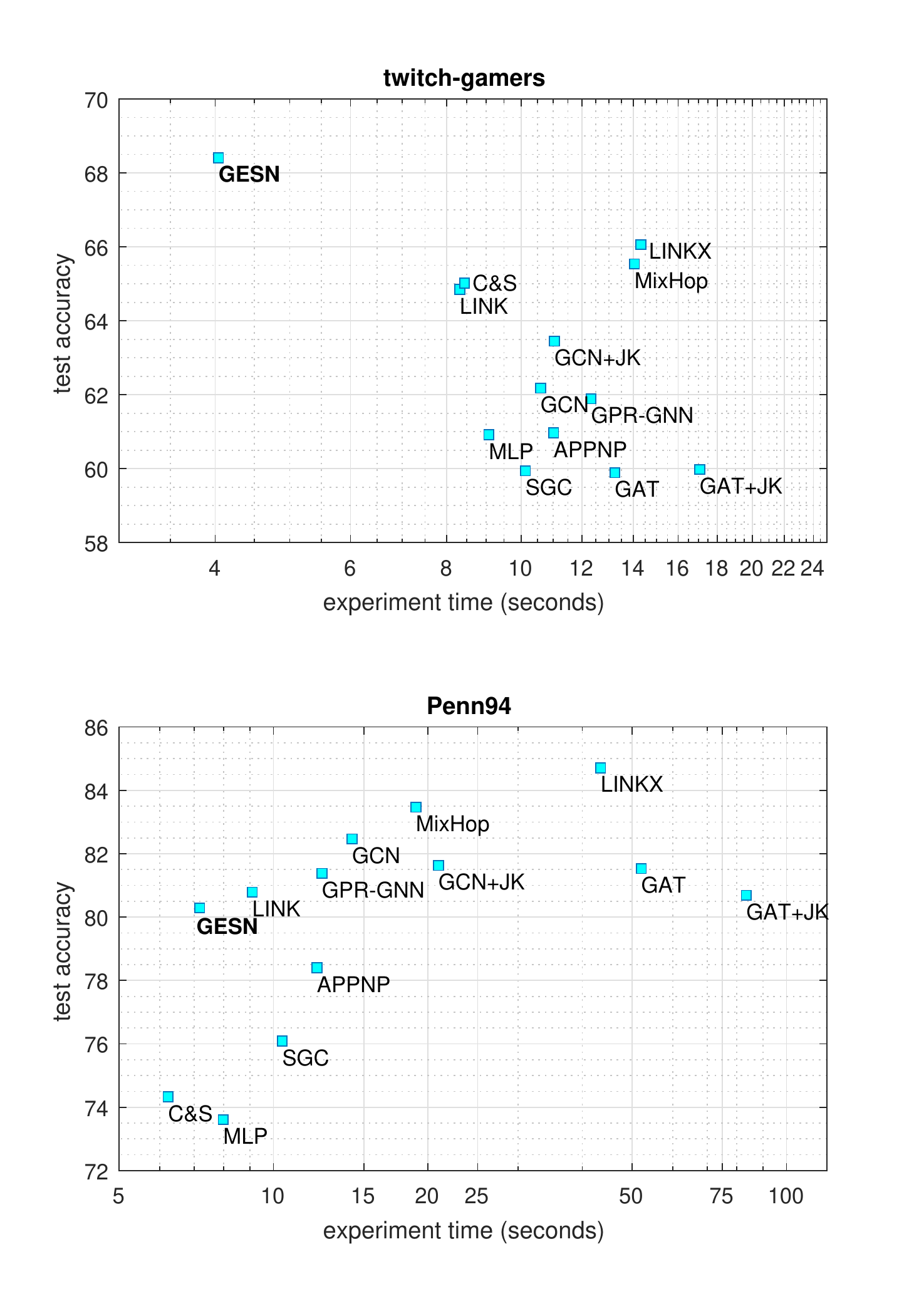}
	\caption{Accuracy/computational time trade-off of GESN and fully-trained models on two large-scale node classification tasks.}
	\label{fig:efficiencty}
\end{figure}

The efficiency/accuracy trade-off for two large-scale tasks is shown in Fig.~\ref{fig:efficiencty}.
For GESN and fully-trained models we measured the time required to perform a training and inference experiment with the largest model configuration (for hyper-parameters of fully-trained models refer to \cite{Lim2021}), averaged over $5$ trials on an Nvidia A100 GPU.
For twitch-gamers, we notice that GESN is from twice to ten times faster with respect to fully-trained models, while offering a gain in accuracy of at least $+2\%$.
GESN does not achieve top accuracy on Penn94, but still offers the same efficiency gain with respect to fully-trained models that have a similar accuracy.
In this task, in particular, the fastest fully-trained models (such as C\&S and MLP) have to significantly sacrifice accuracy, while LINK, as we previously noted, is not an adaptive model, like the top-accuracy LINKX.
We observe that generally the fastest fully-trained models are those that keep the number of trainable weights at minimum, such as C\&S and SGC, or that avoid performing message-passing, such as LINK and MLP.
Apart from GAT and GAT+JK that are particularly demanding due to the attention mechanism, the relative time ranking of fully-trained models varies according to how different number of edges and input feature dimension interacts with the respective model architectures.
Notice that in Fig.~\ref{fig:efficiencty} we adopted $K=100$ iterations for GESN.
As it will be shown in the next section, this number can be greatly tailored down for each task according to the shortest-path distribution of the graph, thus significantly reducing the computation time.
Notwithstanding the much larger embedding dimension required compared to fully-trained models, we can overall conclude that, by avoiding to perform numerous gradient descent epochs for learning node embeddings, GESN can offer an improved efficiency/accuracy trade-off with respect to most fully-trained models.

\section{Analysis}
\label{sec:analysis}

In this section we analyse more in detail different aspects of GESN, with the aim to better understand the motives behind its efficacy.

\subsection{Reservoir radius}

\begin{figure*}
	\centering
	\includegraphics[scale=.55,trim={2.5cm 0 1cm 0},clip]{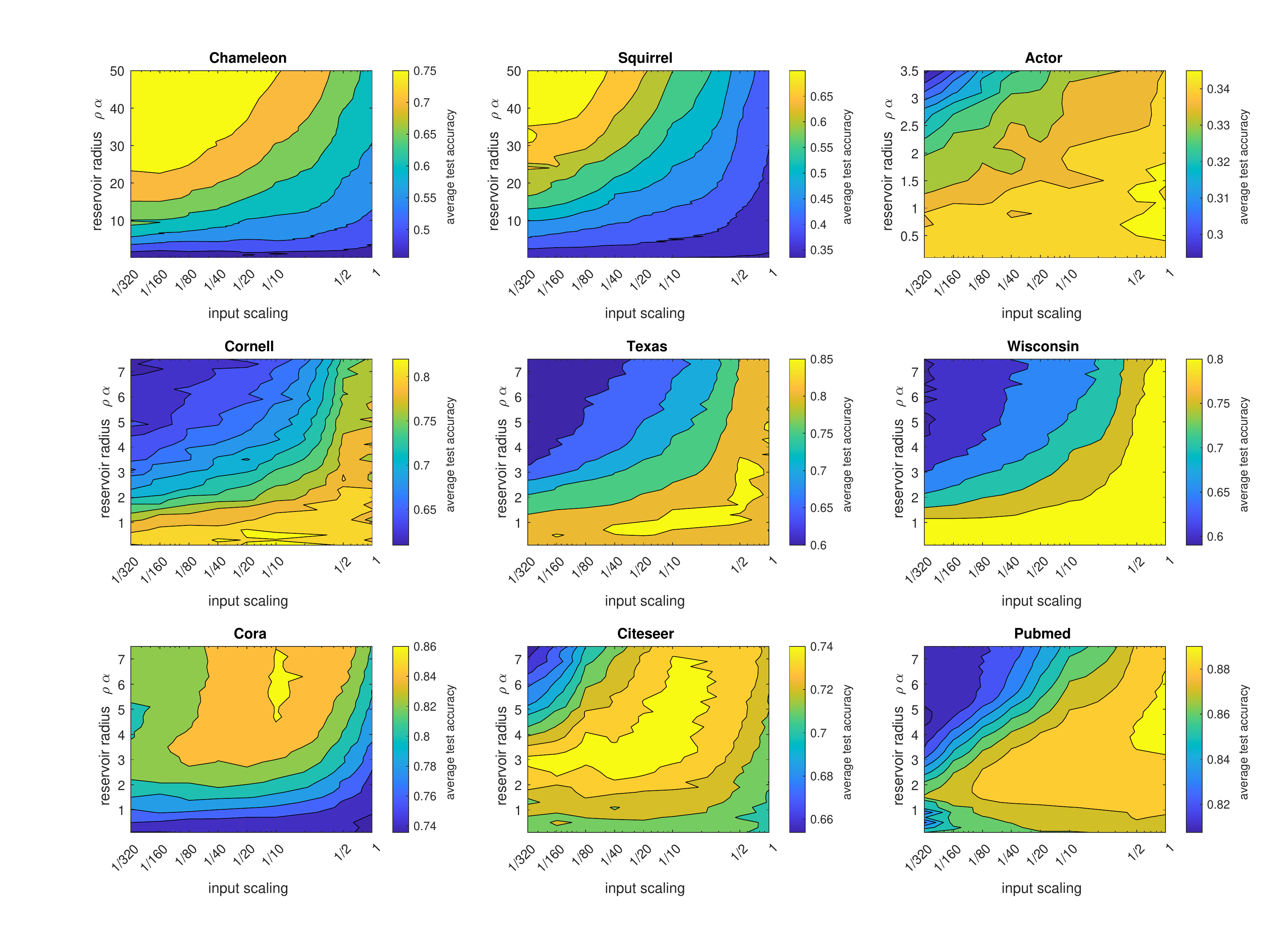}
	\caption{Impact of input scaling and reservoir radius on test accuracy (other parameters set by model selection).}
	\label{fig:rho-scale}
\end{figure*}

In Fig.~\ref{fig:rho-scale}, we show the impact of reservoir radius $\rho$ and input scaling factor on average test accuracy for the small to medium scale tasks.
Chameleon and Squirrel (two tasks with low homophily) require an extremely large reservoir radius, while essentially ignoring the input features due to the extremely small input scaling factor (indeed, removing input features altogether and replacing them with a constant value does not prejudice accuracy).
This indicates that having a non-contractive reservoir with a large Lipschitz constant ($\lVert \mathbf{\hat{W}} \rVert \geq \rho(\mathbf{\hat{W}}) \gg 1$) is beneficial for the extraction of relevant topological features from the graph.
The other four low homophily tasks (Actor, Cornell, Texas, Wisconsin) seem to exploit more the information of node input labels instead of graph connectivity, by requiring reservoir radii within the stability threshold.
Indeed, as noted in section~\ref{sec:gesn}, a contractive reservoir initialization produces an exponentially vanishing sensitivity to more distant node inputs.
Finally, the three high homophily tasks (Cora, Citeseer, Pubmed) achieve the best accuracy with a combination of moderately high spectral radius and input scaling relatively close to $1$.
Overall, what we have observed shows that GESN can be flexible enough to accommodate the two opposite task requirements thanks to the explicit tuning of both input scaling and reservoir radius in the model selection phase.

\subsection{Iterations}

We now analyse the role played by the number of iterations $K$ of the GESN recursive embedding function \eqref{eq:graphesn}, that is the number of message-passing steps.
For a stable reservoir initialization, the convergence to a fixed point is ensured by the contractivity of the state transition function, in turn guaranteed by $\lVert \mathbf{\hat{W}} \rVert < 1$.
In this case, the iterations can be simply allowed to go on until a convergence threshold is reached.
However, the experiments and analysis in this paper have demonstrated that a non-contractive reservoir is crucial for encoding the structural relationships of a node, thus requiring the choice of a suitable value for $K$.

First of all, let us consider a pair of nodes $v, u \in \mathcal{V}$ with their distance, measured as the shortest path length between the two, as $d_{v,u} < \infty$.
Since GESN is performing local aggregations on $1$-hop neighbourhoods, for a `message' from node $v$ to reach node $u$ are required $K > d_{v,u}$ iterations.
Therefore, a $K$ larger than the longest shortest path between all nodes in the graph would allow GESN to process all graph's sub-structures, exploiting them in the node embedding to address the task at hand.

\begin{figure*}
	\centering
	\includegraphics[scale=.6,trim={2cm 0 2cm 0},clip]{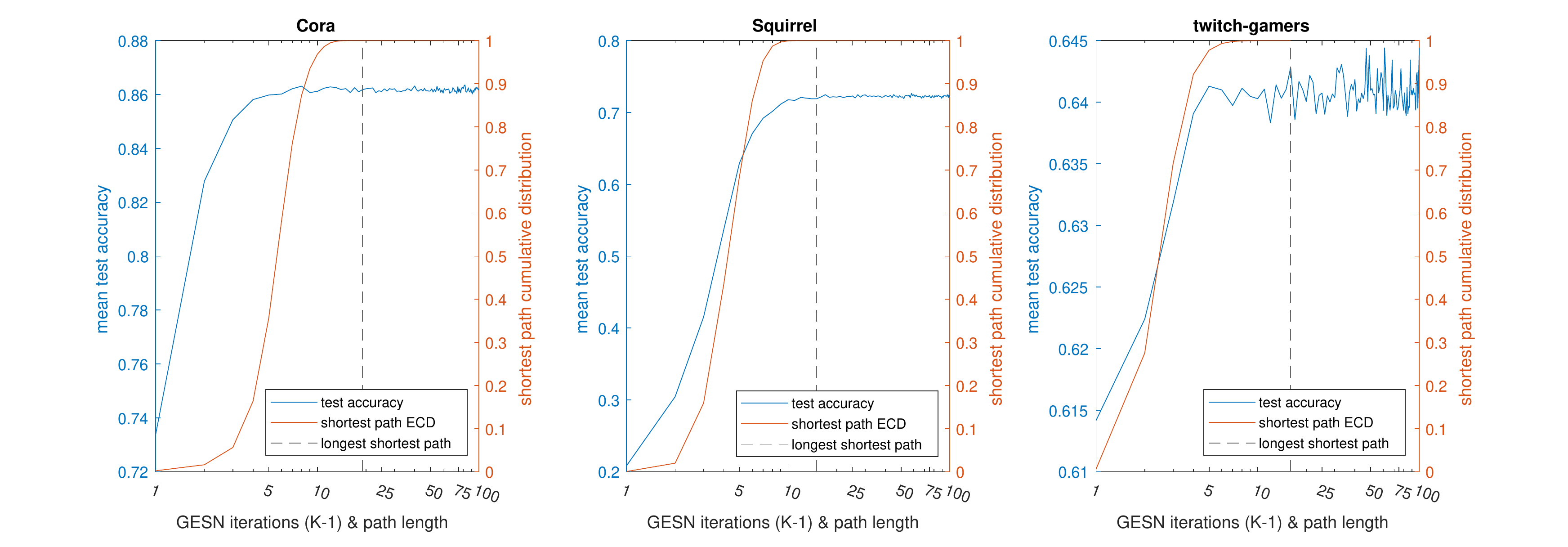}
	\caption{The curve of test accuracy in function of GESN iterations $K-1$ compared against the empirical cumulative distribution (ECD) of shortest paths between reachable nodes in the task graph ($4096$ reservoir units, other parameters chosen by model selection).}
	\label{fig:shortest-paths}
\end{figure*}

In Fig.~\ref{fig:shortest-paths}, we compare the test accuracy for different number of GESN iterations with the cumulative empirical distribution of the shortest paths between connected nodes for three different graphs.
Shortest paths can be computed in $O(|V|^3)$ by the Floyd--Warshall algoritm \cite{Floyd1962} or by specialized versions of the Djikstra's algorithm \cite{Ahuja1990}.
In all three cases, we observe that test accuracy reaches an asymptotic value even before a number of iterations corresponding to the 95th percentile of the shortest path distribution.
In particular, for Cora this asymptotic accuracy is reached already for a smaller number of iterations, around $K = 4$: since this is an homophilic task, information from the closes nodes is already sufficient for providing effective embeddings.
Indeed, the number of GESN iterations is comparable to the number of layers employed in fully-trained convolutional models such as GCN \cite{Li2018}.
In understanding why the accuracy presents an asymptotic behaviour, we can look at the sensitivity equation \eqref{eq:gesn-sensitivity}: the contributions of paths of length $\ell < K$ to the embeddings of node $v$ from a node $u$ at a distance $d_{v,u} < K$ does not diminish as the number of iterations increase.
The same can be intuitively observed in the unrolled representation of GESN in Fig.~\ref{fig:gesn}.
Thus, our analysis offers a simple criterion for selecting $K$ as a parameter: either precisely tailoring it to the specific graph by computing the largest shortest path length, or by setting a value large enough as to capture the whole receptive field of each node, as done in our experiments.

\subsection{Over-smoothing}

\begin{figure*}
	\centering
	\includegraphics[scale=.6,trim={2cm 0 3cm 0},clip]{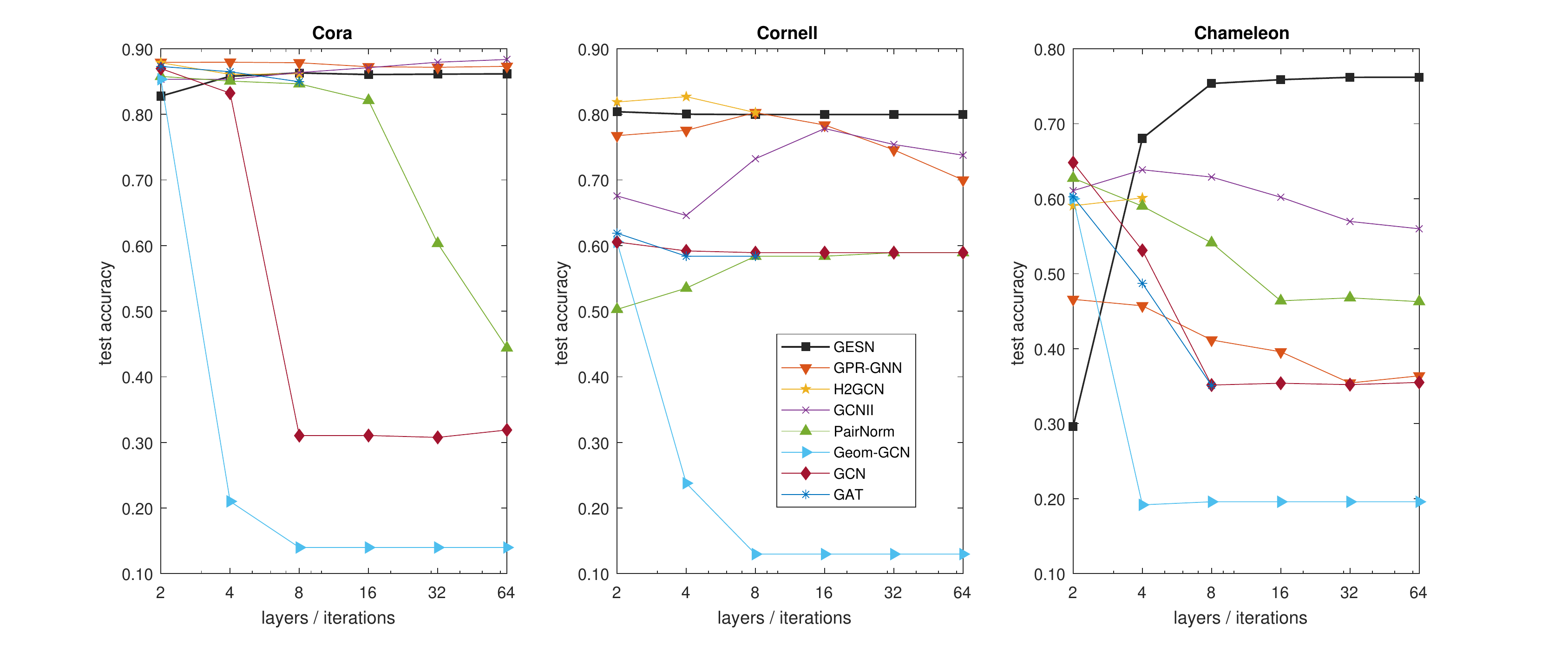}
	\caption{The curves of test accuracy as the number of message-passing layers or iterations increase (except for GESN, values are taken from \cite{Yan2022}).}
	\label{fig:oversmoothing-accuracy}
\end{figure*}

The decline of accuracy as the number of layers increase was the first empirical observation of the over-smoothing problem in deep graph convolutional models \cite{Li2018}.
In GESNs, the iterations of the recursive transition function can be interpreted as equivalent to layers in deep message-passing graph networks where weights are shared among layers, in analogy with the unrolling in RNNs for sequences (Fig.~\ref{fig:gesn}).
As we have just observed in the analysis concerning the number of iterations, this problem seems not to affect GESN.
In Fig.~\ref{fig:oversmoothing-accuracy}, we observe the curve of test accuracy as the number of layers increases in different fully-trained deep convolutional modes and as the number of equivalent iterations increases in GESN.
For convolutional models able to be trained up to $64$ layers, we notice that architectural variations, also employed to ease gradient propagation as in GCNII, can help overcome the phenomenon on the homophilic graph Cora, while an accuracy decline seems nevertheless inevitable on heterophilic graph, in particular on Chameleon.
Noticeably, in the latter GESN presents an inverse trend with respect to fully-trained models, showing an \emph{increase} of test accuracy as a the number of iterations increases.

\begin{figure*}
	\centering
	\includegraphics[scale=.65,trim={3cm 0 3cm 0},clip]{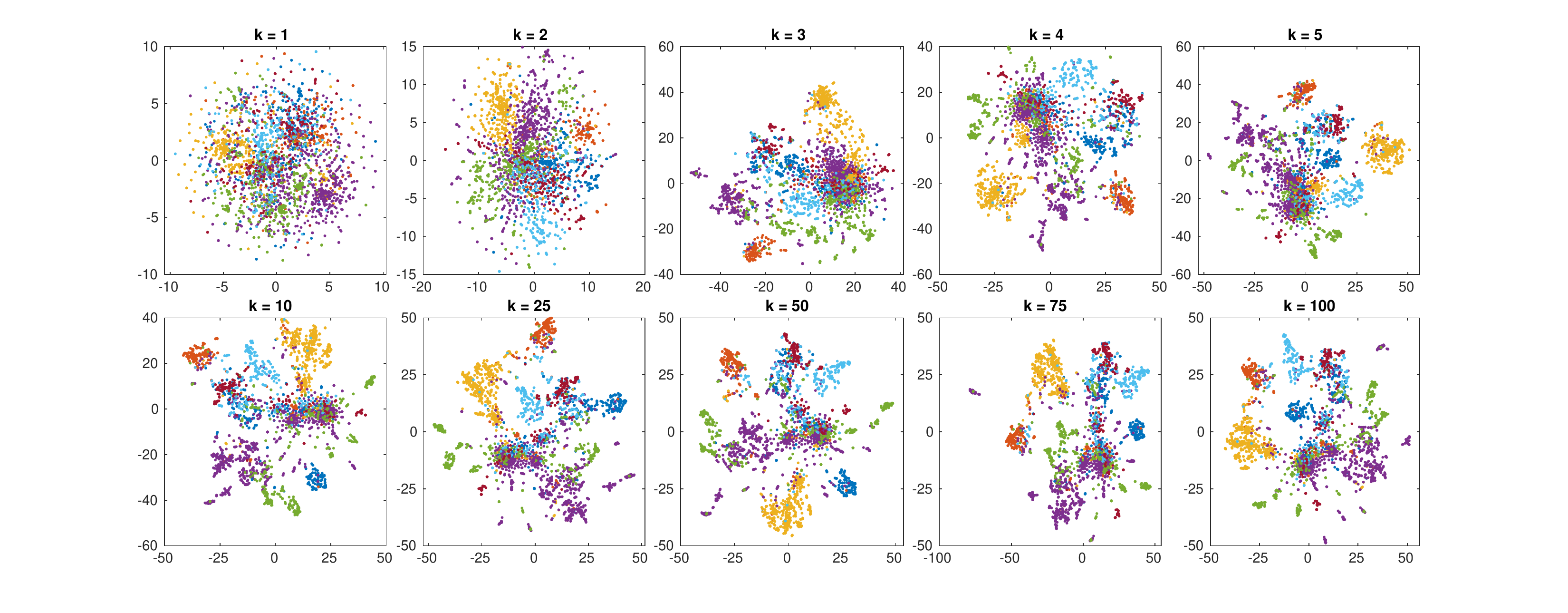}
	\caption{Node embeddings for the Cora graph at different iterations $k$ ($\rho = 6 / \alpha$, $2048$ units). Colours in the t-SNE plots represent different node classes, qualitatively showing how well separable are the node representations. (Best viewed in colour.)}
	\label{fig:cora-iterations}
\end{figure*}

As a further insight, in Fig.~\ref{fig:cora-iterations} we present the t-SNE plots of node embeddings of the Cora graph computed at different iterations of \eqref{eq:graphesn} with reservoir radius set at $\rho = 6 / \alpha$.
We observe that instead of the collapse of node representations that has been shown in \cite{Li2018} and subsequent works on the over-smoothing issue, node embeddings become more and more separable as the number of iterations increases.
This observation, in conjunction with the results of section \ref{sec:experiments}, in particular concerning the comparison against rewiring methods, suggests that the contractivity of the message-passing function, i.e. whether its Lipschitz constant is smaller or larger than $1$, is the critical factor in addressing the degradation of accuracy in deep graph neural networks.
Indeed, tuning the layer contractivity was implicitly done by MADGap \cite{Chen2020} via a regularization term that favours larger pairwise distances of node representations as a mean to address the over-smoothing problem.

\section{Conclusion}
\label{sec:conclusion}

In this paper, we addressed the challenges presented by node classification on heterophilic graphs with GESN, a recursive model based on the reservoir computing paradigm.
Albeit GESN computes the node embeddings without training its weights, our experiments on small- to large-scale graphs have shown that our model can attain accuracy better or in line with fully-trained deep graph message-passing models, while offering a more advantageous trade-off with computational cost.

Our analysis has shown that GESN does not seem to suffer from the over-smoothing and over-squashing issues encountered by other models, as the increase of message-passing steps do not cause a decline in accuracy and no rewiring of the input graph is required to remove message-passing `bottlenecks'.
Instead, our model is able to effectively encode the structural properties of a node, with a number of message-passing iterations dependent on the shortest path distribution in the graph.
In doing so, GESN exhibits an opposite trend with respect to fully-trained deep graph convolutional models, benefiting instead of suffering from an increase in message-passing steps.
A crucial factor is having a non-contractive message-passing function, which our model can explicitly obtain by selecting a sufficiently large reservoir spectral radius.
This is in contrast to previous literature, where convergence to a fixed point of the GESN dynamical system was required to produce effective global graph embeddings.
Our empirical analysis shows that this choice is beneficial for preventing the exponential vanishing of the sensitivity to long-range nodes in the node embeddings.

Future work will involve investigating how the change in Lipschitz constant affects the organization of the node embedding space, and assessing the merit of transferring those results in fully-trained graph convolution models via a regularization term or via constraints on layers' weights.
Furthermore, having an effective and untrained baseline model as GESN offers new opportunities to further the investigation on the issues that plague fully-trained deep graph convolutive models.

\section*{Acknowledgement}
Research partly supported by PNRR - M4C2 - Investimento~1.3, Partenariato Esteso PE00000013 - ``FAIR - Future Artificial Intelligence Research'' - Spoke 1 ``Human-centered AI'', funded by the European Commission under the NextGeneration EU programme.


 \bibliographystyle{elsarticle-num} 
 \bibliography{bibliography}





\end{document}